\def\year{2019}\relax
\documentclass[letterpaper]{article} 
\usepackage{aaai19}  
\usepackage{times}  
\usepackage{helvet}  
\usepackage{courier}  
\usepackage{url}  
\usepackage{graphicx}  
\usepackage{amsmath}
\usepackage{amssymb}
\usepackage{enumerate}
\usepackage{enumitem}
\usepackage{float}

\newcommand{\citenoun}[1]{{\citeauthor{#1} \shortcite{#1}}}
\newcommand{\cut}[1]{}
\newcommand{\x}{\mathbf{x}}
\newcommand{\X}{\mathbf{X}}

\title{When do Words Matter? Understanding the Impact of Lexical Choice on Audience Perception using Individual Treatment Effect Estimation}
\author{Zhao Wang \and Aron Culotta\\
Department of Computer Science\\
Illinois Institute of Technology, Chicago, IL 60616\\
zwang185@hawk.iit.edu, aculotta@iit.edu
}

\begin{document}
%

\maketitle
\begin{abstract}
Studies across many disciplines have shown that lexical choice can affect audience perception. For example, how users describe themselves in a social media profile can affect their perceived socio-economic status. However, we lack general methods for estimating the causal effect of lexical choice on the perception of a specific sentence. While randomized controlled trials may provide good estimates, they do not scale to the potentially millions of comparisons necessary to consider all lexical choices. Instead, in this paper, we first offer two classes of methods to estimate the effect on perception of changing one word to another in a given sentence. The first class of algorithms builds upon quasi-experimental designs to estimate individual treatment effects from observational data. The second class treats treatment effect estimation as a classification problem. We conduct experiments with three data sources (Yelp, Twitter, and Airbnb), finding that the algorithmic estimates align well with those produced by randomized-control trials. Additionally, we find that it is possible to transfer treatment effect classifiers across domains and still maintain high accuracy. 
\end{abstract}

\section{Introduction\footnote{An expanded version of this paper is available at: https://arxiv.org/abs/1811.04890 ; replication files and data are available at: https://github.com/tapilab/aaai-2019-words}}
\label{sec:intro}

Numerous examples from cognitive science, linguistics, and marketing show that lexical choice can affect audience perception~\cite{danescu2012you,ludwig2013more,thibodeau2013natural,riley2015managing,reddy2016obfuscating,preoctiuc2017controlling,packard2017language,nguyen2017fruitful}. For example, a social media user who writes {\it ``I'm excited!"} may be more likely to be perceived as female than one who writes {\it ``I'm stoked!"}~\cite{reddy2016obfuscating}. Similarly, a book with a review containing the sentence {\it ``I loved this book!"} may be perceived as more desirable than one with a review stating {\it ``An excellent novel."}~\cite{ludwig2013more}.

Despite this prior work, we still lack general methods for estimating the causal effect on perception of a single linguistic change in a specific sentence. For example, how much does changing the word {\it ``excited"} to {\it ``stoked"} in the example above increase the chance that a reader will infer the user to be male? Being able to answer such questions has implications not only for marketing and public messaging campaigns, but also for author obfuscation~\cite{hagen2017overview} and stylistic deception detection~\cite{afroz2012detecting}.

A standard empirical approach is to conduct a Randomized Control Trial (RCT), in which subjects are shown texts that differ only in a single linguistic change, and are subsequently asked to rate their perception with respect to a particular attribute. By controlling for the context, we can then attribute changes in perception to the single linguistic change.

Unfortunately, it is impractical to scale such RCTs to the many possible word substitutions across thousands of sentences, making applications based on such methods infeasible. The goal of this paper is to instead investigate automated methods that estimate how a specific lexical choice affects perception of a single sentence. Our approach builds upon a type of causal inference called {\it Individual Treatment Effect} (ITE) estimation. An ITE estimation algorithm estimates the effect of an intervention on an individual; e.g., how effective a drug will be for a specific person. Recently, a number of ITE estimators have been proposed that require only observational data, based on Rubin's potential outcome framework~\cite{rubin1974estimating}. In this paper, we formulate our problem as a type of ITE estimation, which we call {\it Lexical Substitution Effect} (LSE) estimation. We propose two classes of LSE estimators. The first class adapts previous algorithms in ITE estimation to the task of LSE estimation. These methods take as input sentences labeled according to attributes of interest (e.g., a tweet labeled by the gender of the author) and then produces tuples of the form $(w_i, w_j, s, \hat{\tau})$, indicating the estimated LSE ($\hat{\tau}$) of changing the word $w_i$ to $w_j$ for sentence $s$, with respect to the attribute of interest. The second class of estimator is inspired by recent work that frames causal inference as a classification problem~\cite{LopMuaSchTol2015}. This approach requires some labeled examples of the form $(w_i, w_j, s, \tau)$, where $\tau$ is the ``true" LSE according to a RCT. It then fits a classifier based on properties of $(w_i, w_j, s)$ to produce LSE estimates for new sentences.

We conduct studies using three data sources: Airbnb listings, Yelp reviews, and Twitter messages. For Airbnb, we consider the perception of the desirability of a rental based on a sentence from the listing description. For Yelp and Twitter, we consider the perception of the gender of the author. We estimate LSE for thousands of word substitutions across millions of sentences, comparing the results of different LSE estimators. For a sample of sentences, we additionally conduct RCTs using Amazon Mechanical Turk to validate the quality of the algorithmic estimates with respect to human judgments. Overall, we find that the algorithmic estimates align well with those produced by RCTs. We also find that it is possible to transfer treatment effect classifiers across domains and still maintain high quality estimates.

\section{Related Work}
\label{sec:relate}
Studies investigating the effect of wording in communication strategies dates back at least 60 years~\cite{hovland1953communication}. Recent research has explored the effect of wording on Twitter message propagation~\cite{tan+lee+pang:14}, how word choice and sentence structure affects memorability of movie quotes~\cite{danescu2012you}, and how characteristics of news articles influence with high story sharing rates~\cite{Berger@2012}. Additionally, there has been recent psycho-linguistic research discovering how to infer user attributes (e.g., gender, age, occupation) based on language styles. \citenoun{preotiuc2016discovering} explore a wide set of meaningful stylistic paraphrase pairs and verified a number of psycho-linguistic hypotheses about the effect of stylistic phrase choice on human perception. \citenoun{preoctiuc2017controlling} further conduct experiment to control human perception of user trait in tweets. Similarly, \citenoun{reddy2016obfuscating} propose methods to obfuscate gender by lexical substitutions.

As a type of causal inference, individual treatment effect estimation is typically explored in medical trials to estimate effects of drug use on a health outcome. Classical approaches include nearest-neighbor matching, kernel methods and so on \cite{Crump2008nonparametric,lee2008@nonparametric,Willke2012}. However, the performance of these methods do not scale well with the number of covariates \cite{wager2017estimation}. To accommodate a large number of complex covariates, researchers have recently explored  techniques such as random forests \cite{Breiman2001} and causal forests \cite{wager2017estimation}. Motivated by their successful applications in the medical domain, we propose to adapt these techniques to the linguistic domain. Specifically, we conduct experiments to algorithmically estimate the causal effect of lexical change on perception for a single sentence.

In summary, while some prior work has studied overall effects of lexical substitution, in this paper we instead propose methods to estimate context-specific effects. That is, we are interested in quantifying the effect on perception caused by a single word change in a specific sentence. The primary contributions are (1) to formalize the LSE problem as a type of ITE; (2) to adapt ITE methods to the text domain; (3) to develop classifier-based estimators that are able to generalize across domains.

\section{Individual Treatment Effect Estimation}
\label{sec:definition}
In this section, we first provide background on Individual Treatment Effect (ITE) estimation, and then in the following section we will adapt ITE to Lexical Substitution Effect (LSE) estimation.

Assume we have dataset $D$ consisting of $n$ observations $D = \{(\mathbf{X}_1, T_1, Y_1), \ldots, (\mathbf{X}_n, T_n, Y_n)\}$, where $\mathbf{X}_i$ is the {\it covariate vector} for individual $i$, $T_i \in \{0, 1\}$ is a binary {\it treatment} indicator representing whether $i$ is in the treatment ($T_i=1$) or control ($T_i=0$) group, and $Y_i$ is the observed {\it outcome} for individual $i$. For example, in a pharmaceutical setting, $i$ is a patient; $\mathbf{X}_i$ is a vector of the socio-economic variables (e.g., gender, age, height); $T_i$ indicates whether he did ($T_i=1$) or did not ($T_i=0$) receive the medication treatment, and $Y_i \in \{0,1\}$ indicates whether he is healthy ($Y_i=1$) or sick ($Y_i=0$).

We are interested in quantifying the causal effect that treatment $T$ has on the outcome $Y$. The fundamental problem of causal inference is that we can only observe one outcome per individual, either the outcome of an individual receiving a treatment or not. Thus, we do not have direct evidence of what might have happened had we given individual $i$ a different treatment. Rubin's potential outcome framework is a common way to formalize this fundamental problem~\cite{rubin1974estimating}. Let $Y^{(1)}$ indicate the potential outcome an individual would have got had they received treatment $(T=1)$, and similarly let $Y^{(0)}$ indicate the outcome an individual  would have got had they received no treatment ($T=0$). While we cannot observe both $Y^{(1)}$ and $Y^{(0)}$ at the same time, we can now at least formally express several quantities of interest. For example, we are often interested in the {\it average treatment effect} $(\tau)$, which is the expected difference in outcome had one received treatment versus not: $
\tau = \mathop{\mathbb{E}}[Y^{(1)}] - \mathop{\mathbb{E}}[Y^{(0)}]
$. In this paper, we are interested in the {\sl Individual Treatment Effect} (ITE), which is the expected difference in outcome for a specific type of individual:
\begin{align}
\tau(\mathbf{x}) = \mathop{\mathbb{E}}[Y^{(1)} | \mathbf{X}=\mathbf{x}] -   \mathop{\mathbb{E}}[Y^{(0)} | \mathbf{X}=\mathbf{x}]
\label{eq:ite}
\end{align}

that is, the treatment effect for individuals where $\mathbf{X}=\mathbf{x}$. For example, if the covariate vector represents the (age, gender, height) of a person, then the ITE will estimate treatment effects for individuals that match along those variables.

Estimating $\tau({\mathbf{x}})$ from observational data, in which we have no control over the treatment assignment mechanism, is generally intractable due to the many possible confounds that can exist (e.g., patients receiving the drug may be {\it a priori} healthier on average than those not receiving the drug). However, numerous algorithms exist to produce estimates of $\tau({\mathbf{x}})$ from observational data, for example propensity score matching~\cite{austin2008critical}. These methods require additional assumptions, primarily the {\it Strongly Ignorable Treatment Assignment} (SITA) assumption. SITA assumes that the treatment assignment is conditionally independent of the outcome given the covariate variables: $T \perp \{Y^{(0)}, Y^{(1)}\} \mid \mathbf{X}$. While this assumption does not hold generally, methods built on this assumption have often been found to work well. 

With SITA, we can estimate ITE using only observational data as follows:
\begin{align}
\hat{\tau}(\mathbf{x}) = & \mathop{\mathbb{E}}[Y|T=1, \mathbf{X}=\mathbf{x}] - \mathop{\mathbb{E}}[Y|T=0, \mathbf{X}=\mathbf{x}]\\
\label{eq:ite_estimator}
= & \frac{1}{|S_1(\mathbf{x})|} \sum_{i \in S_1(\mathbf{x})} Y_i -\frac{1}{|S_0(\mathbf{x})|}\sum_{i \in S_0(\mathbf{x})} Y_i
\end{align}
where $S_1(\mathbf{x})$ is the set of individuals $i$ such that $\mathbf{X}_i=\mathbf{x}$ and $T_i=1$, and similarly for $S_0(\mathbf{x})$. In other words, Equation~\eqref{eq:ite_estimator} simply computes, for all individuals with covariates equal to $\mathbf{x}$, the difference between the average outcome for individuals in the treatment group and the average outcome for individuals in the control group. For example, if $\mathbf{X=x}$ indicates individuals with (age=10, gender=male, height=5), $T=1$ indicates that an individual receives drug treatment and $T=0$ that they do not, then $\hat{\tau}(\hbox{x})$ is the difference in average outcome between individuals who receive treatment and those who do not.

A key challenge to using Equation~\eqref{eq:ite_estimator} in practice is that $\mathbf{X}$ may be high dimensional, leading to a small sample where $\mathbf{X=x}$. In the extreme case, there may be exactly one instance where $\mathbf{X=x}$. Below, we describe several approaches to address this problem, which we will subsequently apply to LSE estimation tasks.

\section{Lexical Substitution Effect Estimation}
\label{sec:methods}

In this section, we apply concepts from \S\ref{sec:definition} to estimate lexical substitution effect on perception. As a motivating example, consider the following two hypothetical sentences describing the neighborhood of an apartment listed on Airbnb:
\begin{quote}
{\bf A}: There are plenty of {\bf shops} nearby.\\
{\bf B}: There are plenty of {\bf boutiques} nearby.
\end{quote}
We are interested in how substituting {\it shops} with {\it boutiques} affects the perceived desirability of the rental. E.g., because {\it boutiques} connotes a high-end shop, the reader may perceive the rental to be in a better neighborhood, and thus more desirable. Critically, we are interested in the effect of this substitution {\it in one particular sentence}. For example, consider a third sentence:
\begin{quote}
{\bf C}: You can take a 10 minute ride to visit some {\bf shops}.
\end{quote}
We would expect the effect of substituting {\it shops} to {\it boutiques} in sentence {\bf C} to be less than the effect for sentence {\bf A}, since the word {\it shops} in {\bf C} is less immediately associated with the rental.

First of all, to map the notation of \S\ref{sec:definition} to this problem, we specify a sentence to be our primary unit of analysis (i.e., the ``individual"). We make this choice in part for scalability and in part because of our prior expectation on effect sizes --- it seems unlikely that a single word change will have much effect on the perception of a 1,000 word document, but it may affect the perception of a single sentence. The covariate vector ${\bf X}$ represents the other words in a sentence, excluding the one being substituted. E.g., in example sentence {\bf A}, $\mathbf{X} = \langle$ {\it There, are, plenty, of, \_\_, nearby} $\rangle$. 

Second, we note that there are many possible lexical substitutions to consider for each sentence. If we let $p$ index a substitutable word pair ({\it control word $\rightarrow$ treatment word}), then we can specify $T_i^p$ to be the lexical substitution assignment variable for sentence $i$. For example, if $p$ represents the substitutable word pair ({\it shops}, {\it boutiques}), then $T_i^p=0$ indicates that sentence $i$ is in the control group that has the control word {\it shops} in it, and we call it the control sentence, and $T_i^p=1$ indicates that sentence $i$ is treated by substituting the control word {\it shops} to the treatment word {\it boutiques}.

Third, the outcome variable $Y$ indicates the perception with respect to a particular aspect (i.e., desirability or gender in this paper). For example, in the experiments below, we let $Y \in \{1,2,3,4,5\}$ be an ordinal variable expressing the perceived desirability level of an apartment rental based on a single sentence. 

Finally, with these notations, we can then express the {\sl Lexical Substitution Effect} (LSE), which can be understood as the ITE of performing the word substitution indicated by word pair $p$ on a sentence with context words $\mathbf{X}=\mathbf{x}$:
\begin{align}
\tau(\mathbf{x}, p) = \mathop{\mathbb{E}}[Y^{p(1)} | \mathbf{X}=\mathbf{x}] -   \mathop{\mathbb{E}}[Y^{p(0)} | \mathbf{X}=\mathbf{x}]
\end{align}
If we have data of the form $D = \{(\mathbf{X}_1, T_1, Y_1), \ldots, (\mathbf{X}_n, T_n, Y_n)\}$, we can then use the SITA assumption to calculate the LSE: 
\begin{align}
\label{eq:lste_estimator}
\hat{\tau}(\mathbf{x}, p) =  \frac{1}{| S^p_1(\mathbf{x})|}\sum_{i \in S^p_1(\mathbf{x})} Y_i -\frac{1}{| S^p_0(\mathbf{x})|}\sum_{i \in S^p_0(\mathbf{x})} Y_i
\end{align}
where $S^p_1(\mathbf{x})$ is the set of sentences $i$ such that $\mathbf{X}_i=\mathbf{x}$ and $T^p_i=1$, and similarly for $S^p_0(\mathbf{x})$.

As mentioned previously, the high dimensionality of $\mathbf{X}$ is the key problem with using Equation~\eqref{eq:lste_estimator}. This problem is even more critical in the linguistic domain than in traditional ITE studies in clinical domains --- the total number of unique words is likely to be greater than the space of all possible socio-economic variables of a patient. For example, it is entirely possible that exactly one sentence in a dataset has context $\mathbf{X}_i = \mathbf{x}. $\cut{\footnote{Our task is most similar to a personalized medicine task where each user is represented by their genetic profile, which has very high dimension.} }

In the subsections that follow, we first describe four algorithms from ITE estimation literature and how we adapt them to LSE estimation. Then, we describe a classifier-based approach that uses a small amount of labeled data to produce LSE estimates. As a running example, we will consider changing {\it shops} to {\it boutiques} in the sentence {\it ``There are plenty of shops nearby.''} Sentences that contain the control word (e.g., \textit{``shops''}) are called {\sl control samples}, and those containing the treatment word (e.g., \textit{``boutiques''}) are called {\sl treatment samples}. Finally, since we only estimate LSE for one word substitution in one particular sentence each time, we will drop notation $p$ in the following formulas.

\subsection{K-Nearest Neighbor (KNN) Matching}

KNN is a classical approach for non-parametric treatment effect estimation using nearest neighbor voting. The dimensionality problem is addressed by averaging the outcome variables of $K$ closest neighbors. ITE estimation with KNN computes the difference between the average outcome of $K$ nearest neighbors in treatment samples and control samples:
\begin{align}
\hat{\tau}_{KNN}(\x)=\big(\frac{1}{K}\sum_{i \in S_1(\x, K)} {Y_i}\big) - \big(\frac{1}{K}\sum_{i \in S_0(\x, K)} {Y_i}\big)
\label{def.knn}
\end{align}
For an individual with covariate $X=x$, $S_1(\x, K)$ and $S_0(\x, K)$ are the $K$ nearest neighbors in treatment ($T = 1$) and control ($T=0$) samples, respectively.

In LSE estimation with KNN matching, we first represent each sentence using standard tf-idf bag-of-words features, then apply cosine similarity to identify the $K$ closest neighbor-sentences. For our running example, we get $S_0(\x, K)$ by selecting the $K$ sentences that have highest cosine similarity with \textit{``There are plenty of \_\_ nearby''} from the control samples, and get set $S_1(\x, K)$ by selecting the $K$ closest sentences to the treatment samples. Then, the KNN estimator calculates LSE by computing the difference between the average label values of $K$ nearest sentences in the treatment samples and control samples.

\subsection{Virtual Twins Random-Forest (VT-RF)}

The virtual twins approach \cite{Foster@2011subgroup} is a two step procedure. First, it fits a random forest with all observational data (including control samples and treatment samples), where each data is represented by inputs $(\X_i, T_i)$ and outcome $Y_i$. Then, to estimate the ITE, it computes the difference between the predicted values for treatment input $(\X_i, T_i=1)$ and control input $(\X_i, T_i=0)$. The name \textit{`virtual twin'} derives from the fact that for the control input $(\X_i, T_i=0)$, we make a copy $(\X_i, T_i=1)$ as treatment input that is alike in every way to the control input except for the treatment variable. If $\hat{Y}(\x,1)$ is the value predicted by the random forest for input $(\X=\x, T=1)$, then the virtual twin estimate is:
\begin{align}
\label{def.vt}
\hat{\tau}_{VT}(\x)=\hat{Y}(\x,1) - \hat{Y}(\x,0)
\end{align}

where $\hat{Y}(\x,1)$ is the outcome for the \textit{`virtual twin'} (treatment) input and $\hat{Y}(\x,0)$ is the outcome for control input. 

In LSE estimation with VT-RF, we first represent each sentence using binary bag-of-words features (which we found to be more effective than tf-idf). We then fit a random forest to estimate LSE by taking the difference in the posterior probabilities for the virtual twin sentence and the original sentence. For our running example, we fit a random forest classifier using all sentences containing either {\it shops} or {\it boutiques} except the current sentence (for out-of-bag estimation). Meanwhile, we generate the virtual twin sentence{\it ``There are plenty of boutiques nearby.''} Then the estimated LSE is computed by taking the difference between $P(Y=1 |${\it ``There are plenty of boutiques nearby''}$)$ and $P(Y=1 |$ {\it ``There are plenty of shops nearby''}$)$.

\subsection{Counterfactual Random-Forest (CF-RF)}

Counterfactual random forest~\cite{min2018estimating} is similar to VT-RF in that they both calculate ITE by taking the difference between predictions of random forest models. However, CF-RF is different from VT-RF by fitting two separate random forests: a control forest fitted with control samples, and a treatment forest fitted with treatment samples. The ITE is then estimated by taking the difference between the prediction (by treatment forest) for a treatment input ($\hat{Y}_{1}(\x,1)$) and the prediction (by control forest) for a control input($\hat{Y}_{0}(\x,0)$):
\begin{align}
\hat{\tau}_{CF}(\x)=\hat{Y}_{1}(\x,1) - \hat{Y}_{0}(\x,0) 
\end{align}

In LSE estimation with CF-RF, after representing each sentence with binary bag-of-words features, we first fit a control random forest and a treatment random forest and then estimate LSE by taking probability difference between virtual twin sentence and the control sentence. For example, we fit a control forest with all sentences containing {\it shops} excluding the current one (for out-of-bag estimation) and a treatment forest with all sentences containing {\it boutiques}. We then estimate LSE by taking the difference between $P(Y=1 |${\it ``There are plenty of boutiques nearby''}$)$ predicted by treatment forest and $P(Y=1 |${\it ``There are plenty of shops nearby''}$)$ predicted by control forest.

\subsection{Causal Forest (CSF)}

A causal forest~\cite{wager2017estimation} is a recently introduced model for causal estimation. While it also uses random forests, it modifies the node splitting rule to consider treatment heterogeneity. Whereas random forests create splits to maximize the purity of $Y$ labels, causal forests instead create splits by maximizing the variance of estimated treatment effects in each leaf. To estimate ITE for an instance $i$, a causal forest is fit using all treatment and control samples except for instance $i$. Then for each tree in the fitted forest, instance $i$ is placed into its appropriate leaf node in the tree, and the difference between the treated and control outcomes within that node is used as the ITE estimate of that tree. The final estimate is the average estimate of each tree. Let $L(\x)$ be the set of instances in the leaf node to which instance $i$ is assigned, $L_1(\x) \subseteq L(\x)$ be the subset of treatment samples, and $L_0(\x) \subseteq L(\x)$ be the subset of control samples. Then the estimated causal effect of each tree is:
\begin{align}
\hat{\tau}_{CSF}(\x)=\frac{1}{|L_1(\x)|}\sum_{i\in L_1(\x)}{Y_i} - \frac{1}{|L_0(\x)|}\sum_{i \in L_0(\x)}{Y_i}
\end{align}

In LSE estimation with CSF, after representing each sentence with binary bag-of-words features, we fit a causal forest model to estimate LSE by aggregating estimations from all trees. For our running example, we fit a causal forest using all sentences containing either {\it shops} or {\it boutiques}, excluding {\it ``There are plenty of shops nearby''} and then estimate LSE for the sentence by aggregating estimations from all trees, where estimation by each tree is calculated by taking difference between average label values for treatment samples and control samples inside the leaf where {\it ``There are plenty of shops nearby''} belongs to.

\subsection{Causal Perception Classifier}\label{causal_clf}

The advantages of the approaches above is that they do not require any randomized control trials to collect human perception judgments of lexical substitutions. However, in some situations it may be feasible to perform a small number of RCTs to get reliable LSE estimates for a limited number of sentences. For example, as detailed in \S\ref{AMT_LSE}, we can show subjects two versions of the same sentence, one with $w_1$ and one with $w_2$, and elicit perception judgments. We can then aggregate these into LSE estimates. This results in a set of tuples $(w_1, w_2, s, \tau)$, where $\tau$ is the LSE produced by the randomized control trial. In this section, we develop an approach to fit a classifier on such data, then use it to produce LSE estimates for new sentences.

Our approach is to first implement generic, the non-lexicalized features of each $(w_1, w_2, s, \tau)$, then to fit a binary classifier to predict whether a new tuple $(w_1', w_2', s')$ has a positive effect on perception or not. This approach is inspired by recent work that frames causal inference as a classification task~\cite{LopMuaSchTol2015}.

For each training tuple $(w_1, w_2, s, \tau)$, we compute three straightforward features inspired by the intuition of the ITE methods described above. Each feature requires a sentence classifier trained on the class labels (e.g., gender or neighborhood desirability). In our experiments below, we use a logistic regression classifier trained on bag-of-words features.

{\bf 1. Context probability:} The motivation for this feature is that we expect the context in which a word appears to influence its LSE. For example, if a sentence has many indicators that the author is male, then changing a single word may have little effect. In contrast, adding a gender-indicative term to a sentence that otherwise has gender-neutral terms may alter the perception more significantly. To capture this notion, this feature is the posterior probability of the positive class produced by the sentence classifier, using the bag-of-words representation of $s$ {\it after removing word} $w_1$.

{\bf 2. Control word probability:}  This feature is the coefficient for the control word $w_1$ according to the sentence classifier. The intuition is that if the control word is very indicative of the negative class, then modifying it may alter the perception toward the positive class.

{\bf 3. Treatment word probability:}  This feature is the coefficient for the treatment word $w_2$ according to the sentence classifier. The intuition is that if the treatment word is very indicative of the positive class, then modifying the control word to the treatment word may alter the perception toward the positive class.

We fit a binary classifier using these three features. To convert this into a binary decision problem, we label all tuples where $\tau > 0.5$ as positive examples, and the rest as negative.\footnote{We use a 1-5 scale in our RCTs, so a treatment effect greater than 0.5 is likely to be significant.} To compute the LSE estimate for a new tuple $(w_1, w_2, s)$, we use the posterior probability of the positive class according to this classifier.
See detailed analysis in \S\ref{sec:result}.

\section{Data}
\label{sec:data}

This section provides a brief description of experimental datasets (Yelp, Twitter, and Airbnb) for LSE estimation. 

A key benefit of our first class of approaches is that it does not require data annotated with human perceptions. Instead, it only requires objective annotations. For example, annotations may indicate the self-reported gender of an author, or an objective measure of the quality of a neighborhood, but we do not require annotations of user perceptions of text in order to produce LSE estimates. While perception and reality are not equivalent, prior work (e.g., in gender perception from text~\cite{flekova2016analyzing}) have found them to be highly correlated. Our results below comparing with human perception measures also support this notion.

{\bf Neighborhood Desirability in Airbnb:} Airbnb is an online marketplace for short-term rentals, and neighborhood safety is one important factor of desirability that could influence potential guest's decision. Thus, we use crime rate as proxy of neighborhood desirability\cut{\footnote{We only consider Airbnb inside USA, where crime rate is an important factor of neighborhood desirability. This measure might not be suitable for Airbnb out of USA.}}. We collect neighborhood descriptions\footnote{insideairbnb.com} from hosts in 1,259 neighborhoods across 16 US cities and collect FBI crime statistics\footnote{https://ucr.fbi.gov/crime-in-the-u.s/2016}
of each city and crime rate of each neighborhood.\footnote{http://www.areavibes.com/} If a neighborhood has a lower crime rate than its city, we label this neighborhood as desirable; otherwise, undesirable. We get 81,767 neighborhood descriptions from hosts in desirable neighborhoods and 17,853 from undesirable neighborhoods.

{\bf Gender in Twitter Message and Yelp Reviews:} We choose Twitter and Yelp as representative of different social media writing styles to investigate lexical substitution effect on gender perception. First, we use datasets of tweets and Yelp reviews from \cite{reddy2016obfuscating}, where tweets are geo-located in the US and Yelp reviews are originally derived from the Yelp Dataset Challenge released in 2016.\footnote{https://www.yelp.com/dataset\_challenge} Users in both datasets are pre-annotated with male and female genders. \cut{according to Social Security Administration list of baby names\footnote{https://www.ssa.gov/oact/babynames/limits.html}.} In our sample, we have 47,298 female users with 47,297 male users for Twitter dataset, and 21,650 female users with 21,649 male users for Yelp dataset.  Please see Appendix for more details.

\section{Generating Candidate Substitutions}
\label{sec:paraphrase}

Given the combinatorics of generating all possible tuples $(w_1, w_2, s)$ for LSE estimation, we implement several filters to focus our estimates on promising tuples. We summarize these below (see the Appendix for more details): 
\begin{enumerate}
\item Either $w_1$ or $w_2$ must be moderately correlated with the class label (e.g., gender or neighborhood desirability). We implement this by fitting a logistic regression classifier on the labeled data and retaining words whose coefficient has magnitude greater than 0.5.
\item To ensure semantic substitutability, $w_2$ must be a paraphrase of $w_1$, according to the Paraphrase Database (PPDB 2.0) ~\cite{pavlickbetter}.
\item To ensure syntactic substitutability, $w_1$ and $w_2$ must have the same part-of-speech tag, as determined by the most frequently occurring tag for that word in the dataset.
\item To ensure substitutability for a specific sentence, we require that the $n$-grams produced by swapping $w_1$ with $w_2$ occur with sufficient frequency in the corpus.
\end{enumerate}

 After pruning, for Airbnb we obtained 1,678 substitutable word pairs spanning 224,603 sentences from desirable neighborhoods and 49,866 from undesirable neighborhoods; for Twitter we get 1,876 substitutable word pairs spanning 583,982 female sentences and 441,562 male sentences; for Yelp we get 1,648 word pairs spanning 582,792 female sentences and 492,893 male sentences.

\section{Experimental Settings}
\label{sec:experiment}

We first carry out experiments to calculate LSE using four estimators and then conduct Randomized Control Trails with Amazon Mechanical Turk (AMT) workers to get human perceived LSE. Next, we fit an out-of-domain causal perception classifier to distinguish LSE directions. Lastly, we evaluate the performance of each method by comparing with human reported values on each dataset separately.

\subsection{Calculating LSE Estimates}

For experiments with four estimators, we do parameter tuning and algorithm implementation separately. For parameter tuning, we apply the corresponding classification models and do grid search with 5-fold cross validation. For algorithm implementation, we use tuned parameters for each model and follow procedures introduced in \S\ref{sec:methods}.

For KNN, we use KNeighborsClassifier in scikit-learn~\cite{pedregosa2011scikit} for parameter tuning and then select $k=30$ for estimator implementation. For VT-RF and CF-RF, we use RandomForestClassifier (scikit-learn) for parameter tuning and apply the following values in corresponding estimators: $n\_estimators=200$, $max\_features=$`$log2$', $min\_samples\_leaf=10$, $oob\_score=True$. For CausalForest, we use the authors' implementation\footnote{https://github.com/swager/grf} and experiment with $n\_estimators = 200$ and default values for other parameters as suggested by \citenoun{wager2017estimation}. 

For the causal perception classifier, our goal is to determine whether the classifier can generalize across domains. Thus, we train the classifier on two datasets and test on the third.  We use scikit-learn's logistic regression classifier with the default parameters. To compare this classifier with the results of RCTs, we use the posterior probability of the positive class as the estimated treatment effect, and compute the correlation with RCT estimates.

\subsection{Human-derived LSE Estimates} \label{AMT_LSE}

In order to evaluate the methods, and to train the causal perception classifier, we conducted randomized control trails (RCTs) to directly measure how a specific lexical substitution affects reported perceptions. We do so by eliciting perception judgments from AMT workers.

As it would be impractical to conduct AMT for every tuple $(w_1, w_2, s)$, we instead aim to validate a diverse sample of word substitutions rated highly by at least one of the four LSE estimators. For each dataset, we select the top 10 word substitutions that get the highest LSE according to each estimator. For every selected word substitution $(w_1,w_2)$, we sample three control sentences (sentences containing $w_1$) with maximum, minimum and median estimated LSE and generate three corresponding treatment sentences by substituting $w_1$ to $w_2$ for each control sentence. Thus, we get 120 control sentences and 120 treatment sentences for each dataset. \cut{Then we  divide 120 control sentences into 12 batches with each batch has 10 different sentences, and the same process for 120 treatment sentences. } We divide these sentences into batches of size 10; every batch is rated by 10 different AMT workers. The workers are asked to rate each sentence according to its likely perception of an attribute (on a scale from 1 to 5) (e.g., the neighborhood desirability of an Airbnb description sentence, or the gender of author for Twitter and Yelp sentence).  Please see Appendix for details on the annotation guidelines.

For example, for a tuple $(${\it boyfriend}, {\it buddy}, \textit{``My boyfriend is super picky''}$)$, we have 10 different workers rate the likely gender of the author for \textit{``My \underline{boyfriend} is super picky''}, then have 10 {\it different} workers rate the sentence \textit{``My \underline{buddy} is super picky''}. The difference in median rating between the second and first sentence is the human perceived effect of changing the word \textit{boyfriend} to \textit{buddy} in this sentence.

Overall, we recruit 720 different AMT workers, 240 for each dataset, and received 237 valid responses for Yelp, 235 for Twitter, and 215 for Airbnb. We compute the Pearson  correlation between every two workers who rate the same batch as a measure of inter-annotator agreement as well as the difficulty of LSE tasks for each dataset. These agreement measures, shown in Table~\ref{tab.pear_corr}, suggest that the annotators have moderate agreement (.51-.58 correlation) in line with prior work~\cite{preotiuc2016discovering}. Furthermore, these measures indicate that the Airbnb task is more difficult for humans, which is also expected given that neighborhood desirability is a more subjective concept than gender.

\section{Results and Discussion}
\label{sec:result}

In this section, we first show a list of substitution words with large LSE estimates, and then provide quantitative and qualitative analysis for different LSE methods. 

\subsection{Substitution Words with Large LSE}

Table \ref{tab.top_wdpairs} shows a sample of 10 substitution words that have large LSE estimates with respect to desirability or gender, based on the automated methods. For example, replacing \textit{shop} with \textit{boutique} increases the perceived desirability of a neighborhood across many sentences. A sentence using the word \textit{tasty} is perceived as more likely to be written by a male than one using \textit{yummy}, and the word \textit{sweetheart} is more often being used by females than \textit{girlfriend}. 

\begin{table}[t]
    \centering
    \begin{tabular}{ | l | l |  } 
        \hline
        \shortstack{\bf Increase desirability } & \shortstack{\bf Increase male perception } \\
        \hline
        store $\rightarrow$ boutique & gay $\rightarrow$ homo \\ 
        famous $\rightarrow$ grand & yummy $\rightarrow$ tasty \\ 
        famous $\rightarrow$ renowned & happiness $\rightarrow$ joy \\ 
        rapidly $\rightarrow$ quickly & fabulous $\rightarrow$ impressive \\ 
        nice $\rightarrow$ gorgeous & bed $\rightarrow$ crib \\ 
        amazing $\rightarrow$ incredible & amazing $\rightarrow$ impressive \\ 
        events $\rightarrow$ festivals & boyfriends $\rightarrow$ buddies\\
        cheap $\rightarrow$ inexpensive & purse $\rightarrow$ wallet \\
        various $\rightarrow$ several & precious $\rightarrow$ valuable\\
        yummy $\rightarrow$ delicious & sweetheart $\rightarrow$ girlfriend\\
        \hline
        
        \end{tabular}
    \caption{Samples of substitution words with high LSE}
    \label{tab.top_wdpairs}
\end{table}

\subsection{Comparison with RCT Results}

\begin{figure}[t]
	\centering
	\includegraphics[width=\columnwidth]{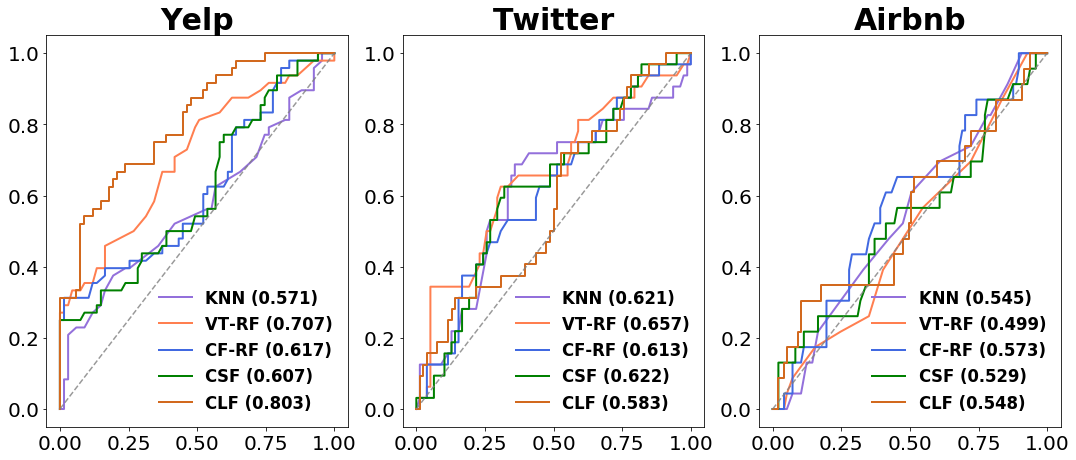}
	\caption{ROC curve for classifying sentences according to AMT perception with estimated LSE as confidence score. {{\it(CLF: causal perception classifier).} Best viewed in color.}}
	\label{fig:auc_cls}
\end{figure}

To evaluate the performance of LSE estimators, we first compare algorithmically derived LSE with human derived LSE from AMT. Each tuple $(w_1, w_2, s)$ has both an algorithmically estimated LSE $\hat{\tau}$ by each estimator as well as a human derived LSE $\tau$ from AMT workers. For 687 annotated tuples, we calculate the Pearson correlation between algorithmic and human-derived LSE estimates. Table \ref{tab.pear_corr} shows the results\cut{\notezw}.{\footnote{Human agreement and algorithmic correlations are calculated differently, so the scores may be in slightly different scales.}} Additionally, Figure \ref{fig:auc_cls} plots ROC curves for classifying sentences as having positive or negative treatment effect, using the LSE estimates as confidence scores for sorting instances.

\begin{table}[t]
    \centering
    \begin{tabular}{ | c | c | c | c |} 
     \hline
     & \textbf{Yelp} & \textbf{Twitter} & \textbf{Airbnb}\\
     \hline
     \hline
     Agreements-pearson & 0.557 & 0.576 & 0.513  \\
    
    \hline
    \hline 
    KNN & 0.474 & 0.291 & 0.076  \\
    VT-RF & 0.747 & 0.333 & 0.049 \\
    CF-RF & 0.680 & 0.279 & 0.109  \\
    CSF & 0.645 & {\bf 0.338} & 0.096 \\
    \hline
    \hline
    Causal perception classifier & {\bf 0.783} & 0.21 & {\bf 0.139} \\ 
    \hline
    \end{tabular}
    \caption{Inter-annotator agreement and Pearson correlation between algorithmically estimated LSE and AMT judgment}
    \label{tab.pear_corr}
\end{table}

\begin{table}[t]
    \centering
    \begin{tabular}{ | r | c | c | c |} 
     \hline
     & \textbf{Yelp} & \textbf{Twitter} & \textbf{Airbnb}\\
     \hline
     {\bf context pr       } & -0.348 & -0.829  & -0.528 \\
     {\bf control word pr  } & -0.141 & -0.514  & -0.367 \\
     {\bf treatment word pr} & ~0.189 & ~0.401  & ~0.344 \\
    \hline
    \end{tabular}
    \caption{Logistic regression coefficients for the features of the causal perception classifier}
    \label{tab.coef2}
\end{table}

From these results, we can see that LSE estimators are well aligned with human perception measures, which suggests the suitable proxy of algorithmic estimators with perception measure. There is also considerable variation across datasets, with Yelp having the most agreement and Airbnb the least. Yelp has the most formal writing style among the three datasets, so tree-based estimators (CF-RF, VT-RF, CSF) have competitive performance with humans. \cut{\notezw}{Twitter is challenging due to grammatical errors and incomplete sentences.} Airbnb has less formal writing style compared with Yelp and contains long sentences with proper nouns (e.g., city names, street names and so on) that lead to the lowest correlation and inter-annotator agreement, suggesting that the more subjective the perceptual attribute is, the lower both human agreement and algorithmic accuracy will be.{\footnote{Using relative crime rates as a proxy for desirability of Airbnb hosts is a possible limitation.}}

Additionally, we observe that the causal perception classifier outperforms the four other LSE estimators for two of the three datasets. Table \ref{tab.coef2} shows coefficient values for the classifier when fit on each dataset separately. These coefficients support the notion that certain aspects of LSE are generalizable across domains --- in all three datasets, the sign and relative order of the coefficients are the same. Furthermore, the coefficients support the intuition as to what instances have large, positive effect sizes: tuples $(w_1, w_2, s)$ where $w_1$ is associated with the negative class (control word probability), where $w_2$ is associated with the positive class (treatment word probability), and where the context is associated with the negative class (context probability).

Finally, we perform an error analysis to identify word pairs for which the sentence context has a meaningful impact on perception estimates. For example, changing the word {\it boyfriend} to {\it buddy} in the sentence {\it ``Monday nights are a night of bonding for me and my \underline{boyfriend}"} is correctly estimated to have a larger effect on gender perception than in the sentence {\it ``If you ask me to hang out with you and your \underline{boyfriend} I will ... decline."} The reason is that the use of the possessive pronoun ``my" reveals more about the possible gender of the author than the pronoun ``your." We found similar results on Airbnb for the words {\it cute} and {\it attractive} --- this change improves perceived desirability more when describing the apartment rather than the owner.

\section{Conclusion}
\label{sec:future}

This paper quantifies the causal effect of lexical change on perception of a specific sentence by adapting concepts from ITE estimation to LSE estimation. We carry out experiments with four estimators (KNN, VT-RF, CF-RF, and CSF) to algorithmically estimate LSE using datasets from three domains (Airbnb, Twitter and Yelp). Additionally, we select a diverse sample to conduct randomized control trails with AMT and fit causal perception classifiers with domain generalizable features. Experiments comparing Pearson correlation show that causal perception classifiers and algorithmically estimated LSE align well with results reported by AMT, which suggests the possibility of applying LSE methods to customize content to perception goals as well as understand self-presentation strategies in online platforms. 

\section*{Acknowledgments}
This research was funded in part by the National Science Foundation under grants \#IIS-1526674 and \#IIS-1618244.

\bibliography{Reference}
\bibliographystyle{aaai}

\clearpage

\appendix
\section{Appendix A: Additional Results}
\label{sec:appa}

We provide supplemental information and detailed analysis in this section.
\subsection{Details for Datasets}
\subsubsection{Airbnb} 
We collect neighborhood descriptions from hosts in 1,259 neighborhoods across 16 US cities from insideairbnb.com by May 2017. Table~\ref{tab.city} shows the 16 cities and corresponding number of neighborhoods we collect in each city.
\begin{table}[H]
    \centering
    \begin{tabular}{ | c | c |} 
     \hline
     \textbf{City} & \textbf{Number of Neighborhoods} \\
     \hline
     LA & 248 \\
     NY & 219  \\
     Oakland & 108  \\
     SanDiego & 96 \\
     Portland & 92 \\
     Seattle & 87 \\
     Denver & 74 \\
     Chicago & 74  \\
     NewOrleans & 69 \\
     Austin & 43 \\
     WDC & 39 \\
     SanFrancisco & 37 \\
     Nashville & 35 \\
     Boston & 25 \\
     Asheville & 8 \\
     SantaCruz & 5 \\
     \hline
    \end{tabular}
    
    \caption{Cities and number of neighborhoods in each city}
    \label{tab.city}
\end{table}

{\bf Neighborhood Desirability} As a subjective concept, desirability of a rental could be measured by multiple factors such as safety, convenience, surroundings, traffic and so on. In this paper, we aim to get an objective measure that could be applied to rentals anywhere and since we only consider Airbnb rentals inside USA, where safety is a very important factor that could influence potential guest's decision, so we decide to use relative crime rate as proxy of neighborhood desirability. However, we acknowledge the limitation of making this assumption. A rental that is attractive to one person who prefer safety might not be attractive to another who prefer location.

We collect crime rate of cities and neighborhoods separately from two sources. For crime rate of cities, we collect from FBI crime statistics\footnote{https://www.freep.com/story/news/2017/09/25/database-2016-fbi-crime-statistics-u-s-city/701445001/}. For crime rate of each neighborhood, we collect from areavibes\footnote{http://www.areavibes.com/}. Considering that crime rate varies from city to city, it is unfair to directly compare neighborhoods in different cities, we make comparisons inside each city by comparing the relative crime rate of a neighborhood with the city it locates as our labeling criteria. We conduct the labeling process as follows: 
\begin{itemize}
  \item Label a neighborhood: if a neighborhood has lower crime rate than the city it locates, we label this neighborhood as desirable; otherwise, undesirable.
  \item Label a host: we assign the same label for hosts located in one neighborhood and get 81,767 neighborhood descriptions from hosts in desirable neighborhoods and 17,853 from undesirable neighborhoods. We observe the data imbalance which might be due to the fact that low-crime areas are more desirable to potential guests, so more Airbnb rentals are listed in low-crime areas than in high-crime areas.
  \item Label a sentence: we label each neighborhood description sentence by the label of that neighborhood, which means all desirable neighborhood description sentences are labeled as desirable, otherwise undesirable.
\end{itemize}

\subsubsection{Twitter and Yelp}
We use tweets and Yelp reviews from datasets introduced in \cite{reddy2016obfuscating}. According to \cite{reddy2016obfuscating}, tweets are collected in July 2013 and only consider those geolocated in US; the corpus of Yelp reviews is obtained from 2016 Yelp Dataset Challenge\footnote{https://www.yelp.com/dataset\_challenge}.  

The two datasets are annotated with two genders: male and female, which are inferred by mapping users' first names with Social Security Administration list of baby names from 1990\footnote{https://www.ssa.gov/oact/babynames/limits.html}. While male and female are suggested as accurate reflection of social media users' genders, we consider non-binary gender labels as an important area of future work. 

After processing by removing users with ambiguous names, dropping non-English and highly gendered texts, they get 432,983 user corpus for Yelp and 945,951 for Twitter \cite{reddy2016obfuscating}. Sampling from their datasets, we get Twitter corpus from 47,298 female users and 47,297 male users, and Yelp corpus from 21,650 female users and 21,649 male users. 

Please refer to \cite{reddy2016obfuscating} for more details about Twitter and Yelp datasets.

\begin{table*}[h]
    \centering
    \begin{tabular}{| p{0.8cm} | p{7cm} | p{7cm} |}
    \hline
      & \textbf{Airbnb} & \textbf{Twitter / Yelp}  \\
     \hline
    Q\&A & Rate the desirability of a short-term apartment rental based on a single sentence. & Rate how likely you think this tweet / Yelp review sentence is written by male or female. \\
    \hline
    5 & Very desirable & \multicolumn{1}{|l|}{Very likely male}\\
    4 & Somewhat desirable & \multicolumn{1}{|l|}{Somewhat likely male} \\
    3 & Neither desirable nor undesirable & \multicolumn{1}{|l|}{Neutral, neither male nor female} \\
    2 & Somewhat undesirable & \multicolumn{1}{|l|}{Somewhat likely female}\\
    1 & Very undesirable & \multicolumn{1}{|l|}{Very likely female}\\
    \hline
    \end{tabular}
    \caption{Amazon Mechanical Turk annotation guidelines}
    \label{tab.amt}
\end{table*}

\subsection{Identify Qualified Lexical Substitutions}
To generate tuples $(w_1, w_2, sentence)$ for LSE estimation tasks, we first search for substitutable word pairs $(w_1,w_2)$ and then select sentences that are qualified for substituting $w_1$ to $w_2$.

\subsubsection{Select Representative Words} Considering the large number of possible lexical substitutions, we first apply several criteria to select the most representative words and then match them with the most appropriate substitutions. To explore the subtle effect of a single word change on perceived perception of the corresponding sentence, we first select words that are representative of attributes we are interested in and thus substituting them might cause effects large enough to be captured. For example, given a sentence: {\it ``I had lunch with my boyfriend''} written by female, {\it boyfriend} is the most representative words with regard to gender of female, substituting {\it boyfriend} to {\it girlfriend} will change the perceived gender of the author from female to male, while substituting the word {\it had} to {\it took} does not change the perceived perception. To select representative words, we fit a binary Logistic Regression classifier for each dataset separately.

For Airbnb dataset, we fit a classifier with 81,767 desirable and 17,853 undesirable neighborhood descriptions. Considering that description texts contain lots of proper nouns like street names, famous place names, neighborhood names and city names, we limit the vocabulary to common words that appear at least 8 times in 6 cities and thus eliminating classifier bias towards proper nouns. By doing so, we get 1,549 common words as representative words of desirable and undesirable classes. 

For Twitter and Yelp datasets, after marking proper nouns with NLTK toolkit\footnote{https://www.nltk.org/}, we fit a binary classifier for Twitter with tweets from 47,298 female users and 47,297 male users. And a classifier for Yelp with reviews from 21,650 female users and 21,649 male users. Using coefficient thresholds greater than 0.5 or smaller than -0.5, we select 4,087 gender representative words for Twitter and 2,264 for Yelp.

After selecting representative words, we search for semantically and syntactically qualified substitutions for them.

\subsubsection{Semantically Qualified Substitutions} \cite{reddy2016obfuscating} apply word2vec extensions of Yelp reviews and tweets parsed with CoreNLP and TweetNLP to capture semantically similar words, and \cite{preotiuc2016discovering} use Paraphrase Database (PPDB) to get stylistic paraphrases with equivalence probability greater than 0.2. In our case, we have three corpus with different writing styles and our goal is to find single word substitutions that express the same meaning, so we choose PPDB as our source to get paraphrases in this paper and will consider word2vec extensions of Airbnb corpus in future work. PPDB(\cite{P15-1146}) is
a collection high precision paraphrases extracted from bilingual parallel corpora with each paraphrase be assigned with probability and similarity scores according to Google ngrams and Gigaword corpus, and later extended with equivalent scores that interpret semantic relationship between paraphrase pairs. We search for paraphrase pairs with equivalence probability of at least 0.15 (\cite{preotiuc2016discovering} use 0.2, we decide to use 0.15 as a relative loose criteria).

\subsubsection{Syntactically Qualified Substitutions} Despite of checking semantics of substitution words, we need to make sure the substitutions are also syntactically qualified. For example, substitutable words should have same singular or plural forms. To do so, we first do POS tagging\footnote{We use NLTK (http://nltk.sourceforge.net/) for POS tagging.} for all sentences in three corpus and store the annotated POS tags of each word, and then check the most common POS tag of each paraphrase pair and only retain paraphrase pairs that have the same most common POS tags.

After limiting substitutions of representative words to semantically and syntactically suitable ones, we search for sentences that are qualified for each specific word substitution.

\subsubsection{Check Word Substitutability in Specific Sentences} We first build a bi-gram vocabulary using three datasets. Then, for each pair of substitution words $(w_1, w_2)$, we search for sentences containing $w_1$ and check for every sentence that if substituting $w_1$ to $w_2$ produces valid bi-grams by looking up the bi-gram vocabulary. For example, to check the substitutability of $(perced, drilled)$ in {\it ``I'm having my ears \underline{pierced} on Saturday''}, we decide the grammatically correctness of the sentence after substitution {\it ``I'm having my ears \underline{drilled} on Saturday''} by checking if {\it ``ears drilled''} and {\it ``drilled on''} exist in our bi-gram vocabulary. If yes, we will keep the current sentence as a qualified sentence for this substitution, otherwise, remove the sentence.

Overall, after pruning with the above criteria, we obtained 1,678 substitutable word pairs spanning 224,603 sentences from desirable neighborhoods and 49,866 from undesirable neighborhoods; and 1,876 substitutable word pairs spanning 583,982 female sentences and 441,562 male sentences for Twitter dataset; and 1,648 word pairs spanning 582,792 female sentences and 492,893 male sentences for Yelp dataset. 

\cut{
\begin{table}[h]
    \centering
    \begin{tabular}{ |p{1.2cm}| p{3cm} |p{3cm} |} 
     \hline
    {\bf Datasets} & {\bf Substitutable Words} & {\bf Qualified sentences}\\
    \hline
    {\bf Yelp} &  & \\
    \hline
    {\bf Twitter} &  & \\
    \hline
    {\bf Airbnb} &  & \\
    \hline
    \end{tabular}
    
    \caption{Final data for LSE tasks}
    \label{tab.LSE_data}
\end{table}
}

\subsection{Crowd-sourcing Experiments with Amazon Mechanical Turk}

We take a tuple $(w_1,w_2,sentence)$ as the unit of analysis in LSE tasks. Despite of algorithmically calculate how much does substituting $w_1$ to $w_2$ for the $sentence$ affects its perceived perception, we conduct Randomized Control Trails to directly measure LSE by eliciting judgments from Amazon Mechanical Turk (AMT) workers. Detailed procedures are as follows:

\begin{itemize}
	\item {\bf Select word pairs with highest LSE} Among all substitution word pairs, we first select those rated highly by at least one of the four LSE estimators (KNN, VT-RF, CT-RF, CSF). Specifically, for each dataset, we get top-10 word substitutions according to each of the four estimators. If a substitution word pair is rated as top-10 with more than one estimators, then we only keep this word pair for the estimator that gives the highest rank (e.g., for a substitution word pair $(w_1,w_2)$, if KNN estimator rank it as the second and VT-RF estimator ranks it as the fifth, then we keep $(w_1,w_2)$ for KNN estimator). Thus, we get 10 substitution word pairs for each of the four estimators.
    \item {\bf Select sentences with maximum, minimum and median LSE for each word pair} For each word substitution $(w_1,w_2)$, we rank all control sentences ({e.g., sentences containing $w_1$}) according to LSE calculated by the corresponding estimator and sample three sentences with maximum, minimum and median LSE. Meanwhile, we generate corresponding treatment sentences using the given substitution word ($w_2$). Thus, we get 120 control sentences and 120 treatment sentences for each dataset. 
    \item {\bf Create AMT tasks} For each dataset, we divide 120 control sentences into 12 batches with each batch has 10 different sentences, and the same process for 120 treatment sentences. We take each batch as a HIT task in AMT, and for each HIT task, we recruit 10 different workers and ask them to pick a scale (ranges from 1 to 5) for every sentence according to its likely perception of an attribute. Table \ref{tab.amt} shows the annotation guidelines for three datasets.
   \item  {\bf Quality control of AMT tasks} To eliminate possible biases, we limit that each worker only have access to one batch of either control or treatment sentences. If a worker rates a batch of control sentences, then he won't be able to see the corresponding treatment sentences, so that his decision is not affected by knowing which word is being substituted. For quality control, we require workers to be graduates of U.S. high schools, and we include attentiveness checks using manually created {\it ``dummy''} sentences. For example, a {\it ``dummy'} sentence for gender perception, \textit{``I am the son of my father''}, should be rated as written by a male. We remove responses from workers who provide incorrect answers for dummy questions.     
\cut{    \item  We design {\it ``dummy''} sentences for attentiveness checks of AMT workers as shown in Table~\ref{tab.dummy}. To eliminate bias, we mix dummy sentences with sentences chosen for LSE RCTs. }
\end{itemize}
 
\begin{table}[h]
    \centering
    \begin{tabular}{ | p{6.6cm} |p{1.52cm} |} 
     \hline
    {\textbf{Yelp }} & {\bf Label}\\
    \hline
    My wife likes this place. & Male\\ I like coming here with my fraternity brothers. & Male\\ My brother and I come here for guys night out. & Male\\ My husband likes this place. & Female\\ I like coming here with my sorority sisters. & Female\\ My sister and I come here for girl's night out. & Female \\
    \hline \hline
    {\textbf{Twitter }} & \\
    \hline
    I love playing football and video games. &  Male\\ My wife is waiting on me. & Male\\ I am my father's son. & Male\\ I love getting a pedicure at girls night out. & Female\\ My husband says I smile too much. & Female\\ I am my mom's daughter. & Female \\
    \hline \hline
     {\textbf{Airbnb }} & \\
    \hline
    This is by far the best neighborhood in the city. & Desirable \\ This neighborhood is amazing in every way. & Desirable\\ What a world-class neighborhood this is! & Desirable\\ This neighborhood is not so great. & Undesirable\\ Yes, there is a lot of crime in this neighborhood. & Undesirable\\ Lots of shootings in this neighborhood. & Undesirable \\
    \hline
    \end{tabular}
    
    \caption{Dummy sentences for Yelp, Twitter and Airbnb}
    \label{tab.dummy}
\end{table}

\subsection{Experiments with LSE Estimators}
We first conduct parameter tuning to select the most suitable parameters for each estimator and then implement four estimators following procedures introduced in \S\ref{sec:methods}.

\subsubsection{Parameter Tuning}
As we are estimating LSE on sentence level, we do parameter tuning with all labeled sentences of each dataset. Parameters are tuned for the classification task, but not for the treatment effect estimation task (none of the KNN/VT-RF/CF-RF/CSF methods were tuned using the labeled AMT data, so we can measure effectiveness without access to such expensive data).

\begin{itemize}
	\item {\bf Feature Representation} We try both bag-of-words and tf-idf feature representation techniques for each method.
    \item {\bf KNN tuning} We use scikit-learn implementation of KNeighborsClassifier and do grid search for $n\_neighbors$ (since we only need the number of neighbors in KNN estimator implementation, so we don't consider other parameters) and get the best 5\-fold cross validation score with $ n\_neighbors = 30$.
	\item {\bf Random-Forest tuning} We use scikit-learn implementation of Random Forest classifier and do grid search for a set of parameters and get the best 5-fold cross validation score with $n\_estimators = 200$, $max\_features = ${\it `log2'}, $min\_samples\_leaf = 10$ and $oob\_score = True$.
\end{itemize}

As mentioned in previous context, there exists imbalance between the number of `desirable' and `undesirable' descriptions in Airbnb dataset. We considered model variants that deal with class imbalance (e.g., overweighting the minority class), but did not observe significantly different results with such methods.

\subsubsection{Estimator Implementation}
For estimator implementation, we follow the process introduced in \S\ref{sec:methods} and use the best parameters reported by the above tuning process for KNN VT-RF, CF-RF. For Causal Forest, we try $n\_estimators=200$ with default values of other parameters.

\subsection{Causal Perception Classifier}
We fit two classifiers for this task. First, we fit one classifier for each dataset to get proposed features: posterior probability of a context, coefficient of substitution words, and the number of positively and negatively related words. After representing each tuple $(w_1,w_2,sentence)$ with proposed features, we fit  causal perception classifiers only using samples labeled by Amazon Mechanical Turks. Specifically, each causal perception classifier is fitted by using samples of two datasets and making out-of-domain prediction for the third dataset.

\begin{table*}[h]
    \centering
    \begin{tabular}{ | l | l | l |  } 
        \hline
        \textbf{Yelp} & \textbf{Twitter} & \textbf{Airbnb} \\
        \hline
        lovely $\rightarrow$  delightful & gay $\rightarrow$ homo & store $\rightarrow$ boutique \\ 
        cute $\rightarrow$ attractive & yummy $\rightarrow$ tasty & famous $\rightarrow$ grand \\ 
        helpful $\rightarrow$ useful & happiness $\rightarrow$ joy & famous $\rightarrow$ renowned\\ 
        fabulous $\rightarrow$ terrific & fabulous $\rightarrow$ impressive & rapidly $\rightarrow$ quickly\\ 
        gorgeous $\rightarrow$ outstanding & bed $\rightarrow$ crib & nice $\rightarrow$ gorgeous \\ 
        salesperson $\rightarrow$ dealer & amazing $\rightarrow$ impressive & amazing $\rightarrow$ incredible\\ 
        belongings $\rightarrow$ properties & boyfriends $\rightarrow$ buddies & events $\rightarrow$ festivals  \\
        thorough $\rightarrow$ meticulous & purse $\rightarrow$ wallet & cheap $\rightarrow$ inexpensive\\
        happily $\rightarrow$ fortunately & precious $\rightarrow$ valuable & various $\rightarrow$ several \\
        dirty $\rightarrow$ shitty & sweetheart $\rightarrow$ girlfriend & yummy $\rightarrow$ delicious\\
        \hline
        \multicolumn{2}{|c|}{Increase male perception or decrease female perception} & Increase desirability \\
        \hline
        \end{tabular}
    \caption{Substitutable word pairs with large LSE}
    \label{tab.top10_wdpairs}
\end{table*}

\subsection{Results and Analysis}
In this section, we provide both qualitative and quantitative analysis from the following aspects:
\begin{itemize}
    \item First, we present a sample of substitution words estimated to have large LSE.\cut{ and a subjective comparison of top rated substitutions according to each estimator.}
    \item Second, we compare the performance of four LSE estimators.
    \item Third, we evaluate the agreement of each estimator with human perception RCTs using Amazon Mechanical Turk.
    \item Fourth, we assess the causal perception classifier and interpret feature importance with experimental findings.
    \item Finally, we provide a preliminary analysis of how this approach may be used to characterize communication strategies online.
\end{itemize}

\subsubsection{Substitution Words with Large LSE}

Table~\ref{tab.top10_wdpairs} shows a sample of substitutable word pairs estimated to have large LSE by at least one estimator. 

For Airbnb, the substitution words are reported to increase the perceived desirability of a rental. For example, since {\it boutique} often related with nice neighborhoods, substituting {\it shop} to {\it boutique} helps increase the neighborhood desirability. For Twitter and Yelp, the substitution words are reported to increase male perception or decrease female perception of the author. For example, a sentence using {\it tasty} is more likely to be written by a male than using {\it yummy}, and chances are high that {\it sweetheart} would appear in a female sentence while {\it girlfriend} in a male sentence.

Additionally, to assess the quality of substitutable word pairs, we select top 20 word pairs with largest LSE reported by each estimator and manually check if these word pairs are both syntactically and semantically qualified substitutions. As indicated by Table~\ref{tab.top20_accuracy}, we find that KNN estimator is somewhat more likely to assign large LSE for qualified substitutions.\cut{ We suspect this is in part due to the fact that KNN estimator tends to assign large LSE when it is able to identify control and treatment sentences that are highly similar. In these cases, the words tend to be substitutable.} Unsuitable word pairs are often generated due to the fact that the paraphrase database (PPDB) was trained on general texts, but the validity of a substitution can depend on domain. For example, {\it gross} and {\it overall} are potential paraphrases according to PPDB due to one sense of {\it gross}, but in the Twitter data {\it gross} is instead more commonly used as a synonym for {\it disgusting}. More conservative pruning using language models trained on the in-domain data may reduce the frequency of such occurrences.

\begin{table}[t!]
    \centering
    \begin{tabular}{ | c | c | c | c | c |} 
     \hline
      & \textbf{Yelp} & \textbf{Twitter} & \textbf{Airbnb} & \textbf{Mean} \\
     \hline
    KNN & 100\%  & 85\%  & 90\% & {\bf 91.67\%} \\    
    VT-RF & 100\% & 65\% & 90\% & 85\% \\
    CF-RF & 85\% & 75\% & 75\% & 78.33\%  \\
    CSF & 80\% & 70\% & 50\% & 66.67\%  \\
    \hline
    \end{tabular}
    
    \caption{Fraction of top 20 substitutable word pairs that are judged to be acceptable by manual review}
    \label{tab.top20_accuracy}
\end{table}

\subsubsection{Quantitative Analysis of LSE Estimators}

In this section, we quantitatively compare the similarities and differences between four estimators. We expect there to be differences between KNN and the forest-based methods, since their underlying classification functions are different: KNN estimator directly search from all training instances to identify $k$ nearest neighbors in control and treatment group. In contrast, VT-RF, CF-RF and CSF are all tree-based methods, which attempt to place instances in the same leaf if they are homogeneous with resepect to the covariate vector $\X$.

To quantitatively compare the performance of four estimators, we first generate the entire ranked list of $(w_1,w_2,sentence)$ tuples according to each estimator and then compute Spearman's rank correlation for ranked list of every two estimators.

\begin{table*}[h!]
    \centering
    \begin{tabular}{ |c|c|c|c|c|c|c|c|c|c|c|c|c|} 
     \hline
      & \multicolumn{4}{| c }{\textbf{Yelp}} & \multicolumn{4}{| c |}{\textbf{Twitter}} & \multicolumn{4}{| c |}{\textbf{Airbnb}} \\
     \hline
       & KNN & VT-RF & CF-RF & CSF & KNN & VT-RF & CF-RF & CSF & KNN & VT-RF & CF-RF & CSF\\
     \hline
    KNN & 1.0 & 0.674 & 0.715 & 0.655 & 1.0 & 0.699 & 0.729 & 0.668 & 1.0 & 0.469 & 0.561 & 0.455 \\
    VT-RF & & 1.0 & 934 & {\bf 0.945} & & 1.0 & 0.932 & \textbf{0.935} & & 1.0 & {\bf 0.822} & 0.773 \\
    CF-RF & & & 1.0 & 0.899 & & & 1.0 & 0.883 & & & 1.0 & 0.733 \\
    CSF & & & & 1.0 & & & & 1.0 & & & & 1.0 \\
    \hline
    \end{tabular}
    
    \caption{Spearmanr correlation between ranked sentences of four estimators}
    \label{tab.sent_corr}
\end{table*}

According to results shown in Table ~\ref{tab.sent_corr}, we observe that: 
\begin{itemize}
	\item Forest based methods (VT-RF, CF-RF, CSF) perform more similar than KNN. 
	\item Four estimators have less agreement on Airbnb dataset than on Twitter and Yelp, which suggests that estimating LSE on Airbnb is harder, because hosts are incentivized to highlight desirable aspects of the neighborhood.
\end{itemize}

Then, we calculate the percentage of sentences labeled as negative (refers to undesirable for Airbnb and female for Yelp and Twitter) among top 1000 sentences with large LSE. Results in Table~\ref{tab.neg_percentage} shows that: 
\begin{itemize}
	\item All of the four estimators tend to pick negative instances for large LSE. Since we rank sentences in descending order of estimated LSE, the more number of negative sentences ranked in top 1000 the more effective that estimator is.
   \item CF-RF estimator picks the most negative instances for large LSE.
   \item VT-RF estimator performs differently with other estimators, and especially for Airbnb dataset. The reason may lie in the fact that we label each description sentences as desirable or undesirable according to relative crime rate of a neighborhood, which means all sentences describing low-crime neighborhoods are labeled as desirable and vice versa. However, sentences describing low-crime neighborhoods are not guaranteed to disclose desirability but will be mislabeled as desirable according to our criteria, and this misleads VT-RF estimator and explains the difference of this estimator.
\end{itemize}

\begin{table}[H]
    \centering
    \begin{tabular}{ |c|c|c|c|} 
     \hline
     & \textbf{Yelp} & \textbf{Twitter} & \textbf{Airbnb} \\
     \hline
     KNN & 90.5\% & 70.5\% & 86.6\% \\
     VT-RF & 93.9\% & 71.3\% & 64.9\% \\
     CF-RF & \textbf{96.6\%} & \textbf{77.3\%} & \textbf{87.7\%} \\
     CSF & 95.9\% & 71\% & 84.4\% \\
    \hline
    \end{tabular}
    \caption{Percentage of negative sentences in top 1000 highly ranked instances with respect to LSE}
    \label{tab.neg_percentage}
\end{table}


\subsubsection{Qualitative Analysis of Estimators}

To qualitatively assess the performance four estimators, we first show examples to get a better understanding of how do four estimators perform differently in recommending substitutable words for a sentence. As shown in Table ~\ref{tab.difword_samesents}: 
\begin{itemize}
	\item For Yelp, we pick a sentence labeled as male and find substitutable words to make it more likely a sentence written by female. Four estimators give same recommendations for this sentence.
    \item For Twitter, we pick a sentence labeled as female and four methods recommend substitutable words to make it more likely a male sentence. E.g., as {\it boyfriend} is most likely to be used by females while {\it buddy} by males, substituting {\it boyfriend} to {\it buddy} makes the sentence more likely to be perceived as written by male. 
    \item For Airbnb, we pick a neighborhood description sentence labeled as undesirable, and four estimators make recommendations to improve its desirability. CF-RF and CSF agree on recommendations for this sentence.
\end{itemize}
Additionally, we show an example in table~\ref{tab.sameword_difsents} to see how do LSE vary for same word substitution in different sentences. We randomly pick one substitutable word pair in each dataset, and get its highest and lowest LSE sentence according to CSF estimator. 
\begin{itemize}
	\item For Airbnb, substituting {\it shop} to {\it boutique} gives lowest LSE on the sentence that is less immediately associated with rental, because it is {\it ``located a mile away''}.
    \item For Twitter, substituting {\it boyfriend} to {\it buddy} gives highest treatment effect for the sentence talking about {\it ``my boyfriend''}, which the word {\it ``my''} is directly associated with the writer of this sentence, so substituting it to {\it ``my buddy''} makes a big change on the writer's gender. But for the lowest treatment sentence, the substitution makes a small change because {\it ``your boyfriend''} and {\it ``your buddy''} do not refer to the writer's gender.
\end{itemize}
 \begin{figure*}[h]
	\centering
	\includegraphics[width=2\columnwidth]{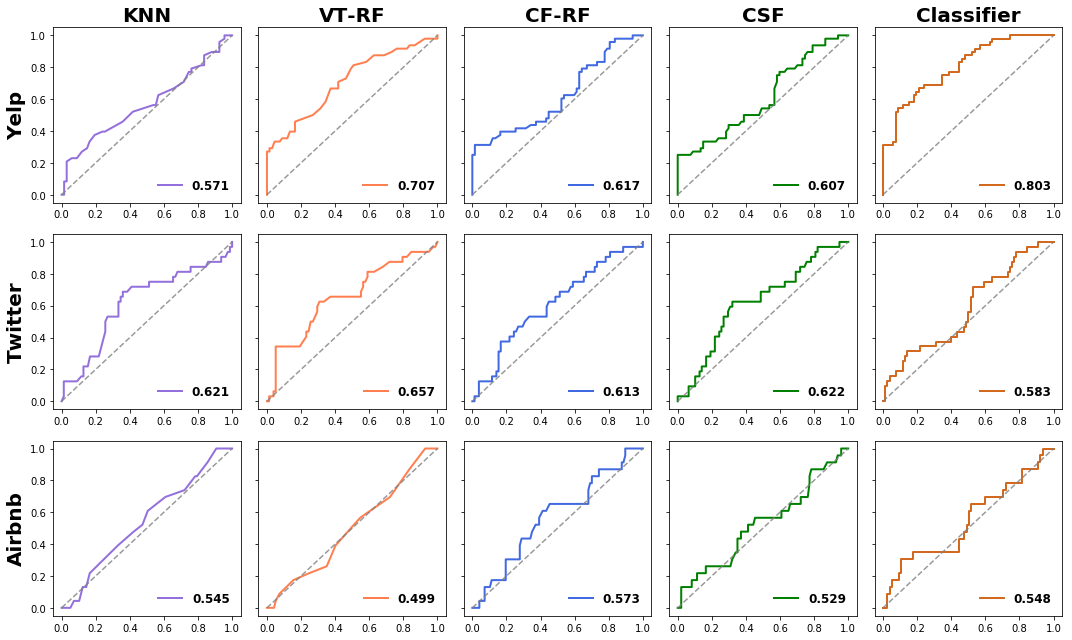}
	\caption{ROC curve for classifying sentences according to AMT perception with estimated LSE as confidence score }
	\label{fig:apd_auc_cls}
\end{figure*}

     
    
    
    

 \subsubsection{Performance of Causal Perception Classifier}
Our goal for causal perception classifier was to use a small number of generic features to allow the method to generalize across domains (e.g., we fit a model on Yelp and apply it to Twitter). Despite of results shown in  \S\ref{sec:result}, we performed some preliminary experiments with a few other features (e.g., sentence length, part-of-speech, number of support words and conflict words), but did not observe significantly different results.
 
 \begin{table}[H]
    \centering 
    \begin{tabular}{ | l | c | c | c |} 
     \hline
     \textbf{Performance} & \textbf{Yelp} & \textbf{Twitter} & \textbf{Airbnb}  \\
     \hline
    AUC & 0.803 & 0.583 & 0.548  \\
    Precision & 0.80 & 0.70 & 0.65 \\
    Recall & 0.69 & 0.72 & 0.81 \\
    F1 & 0.63 & 0.62 & 0.72 \\
    \hline
    \end{tabular}
    \caption{Performance of causal perception classifier}
    \label{tab.clf}
\end{table}

\cut{
\begin{table}[H]
    \centering
    \begin{tabular}{ | c | c | c | c |} 
     \hline
     & \textbf{Yelp} & \textbf{Twitter} & \textbf{Airbnb}\\
     \hline
     \hline
     Agreements-pearson & 0.557 & 0.576 & 0.513  \\
    
    \hline
    \hline 
    KNN & 0.474 & 0.291 & 0.076  \\
    VT-RF & 0.747 & 0.333 & 0.049 \\
    CF-RF & 0.680 & 0.279 & 0.109  \\
    CSF & 0.645 & {\bf 0.338} & 0.096 \\
    \hline
    \hline
    Causal perception classifier & {\bf 0.783} & 0.21 & {\bf 0.139} \\ 
    \hline
    \end{tabular}
    \caption{Inter-annotator agreement and Pearson correlation between algorithmically estimated LSE and AMT judgment}
    \label{tab.corr}
\end{table}
}
\subsubsection{Comparing LSE Reported by Estimators with Human Judgments}
In this section, we evaluate the agreement between LSE estimators with human perception RCTs using Amazon Mechanical Turk. To do this, we first calculate inner-annotator agreement using both pearson and Spearman's rank and take it as a measure of the difficulty of LSE task with each dataset, and then compute Pearson correlation between LSE reported by four estimators. For the RCTs, we compute human perceived LSE as the difference between median ratings for treatment sentence and control sentence.

\cut{As a problem addressed in \S\ref{sec:methods}, it is rare to find text datasets annotated with perception. In this paper, we first experiment with data labeled by objective measures. For Airbnb, we take crime rate as an objective measure of the quality of a neighborhood. For Yelp and Twitter, we take the self-reported gender as an objective measure. Then, we compare results from objective measures with perception measures from AMT. }

Table~\ref{tab.pear_corr} shows the Pearson correlation between each LSE estimator and AMT reported LSE and Figure~\ref{fig:apd_auc_cls} shows the extend ROC curve for classifying sentences according to AMT perception with estimated LSE as confidence score.
\begin{itemize}
	\item LSE estimated by four estimators are well aligned with AMT perceived results, which suggests the suitable proxy of objectives measures we use with perception measure. Specially treatment effect for Yelp dataset calculated by CF-RF method has the highest correlation 0.57. 
    \item LSE task for Yelp has the highest correlation with AMT perceived results and three tree-based methods (CF-RF, VT-RF, CSF) have competing performance with innter-annotator agreement.
    \item LSE task for Airbnb has the lowest correlation, and inter-annotator agreement by pearson and Spearman's rank also be the lowest, which suggests the difficulty for LSE estimation on Airbnb dataset. 
    \item The more subjective the perceptual attribute is, the lower both human agreement and machine accuracy will be. Since Yelp has the most formal writing style among the three datasets, LSE estimators perform as good as humans. Twitter is challenging due to its informal writing, as compared with Yelp, which contains more grammatically correct, complete sentences. Beyond the fact that desirability is subjective, Airbnb has an informal writing style and contains long sentences with proper nouns (e.g., city names, street names and so on), which decrease the sentence readability for AMT workers. Besides that, since hosts are motived to attract guests by highlighting positive aspects and roundabout negative aspects, the use of euphemism increases the difficulty of this task: on one hand, it increases the difficulty for human understanding; on the other hand, it misleads LSE estimators as we did not equip LSE algorithms with the ability to identify euphemisms.
    \item Estimators' performance with Yelp dataset correlate with humans more than humans correlate with each other. We have two possible explanations for this: First, human agreement is calculated from the average pairwise correlation across 10 AMT workers annotating the same 10 sentences. In contrast, the algorithmic correlations are calculated by comparing the algorithmic scores with the median human scores across 200 or so sentences. Because of this somewhat different calculation, the scores may be in slightly different scales. Second, while we implemented several quality control measures for AMT (see the end of section A.3 in the supplementary material), there are still some outlier workers who reduce the overall agreement number. This in part motivates our use of the median rating when computing the final results.
 
\end{itemize}

\cut{
\begin{table}[t]
    \centering
    \begin{tabular}{ | r | c | c | c |} 
     \hline
     & \textbf{Yelp} & \textbf{Twitter} & \textbf{Airbnb}\\
     \hline
     {\bf context pr       } & -0.348 & -0.829  & -0.528 \\
     {\bf control word pr  } & -0.141 & -0.514  & -0.367 \\
     {\bf treatment word pr} & ~0.189 & ~0.401  & ~0.344 \\
    \hline
    \end{tabular}
    \caption{Logistic regression coefficients for the features of the causal perception classifier.}
    \label{tab.coef}
\end{table}
}

In addition to correlation, we also evaluate whether the sign of algorithmically estimated LSE agree with AMT perceived LSE. To do so, we code estimated LSE as positive or negative, and compute ROC curves for each estimator shown in Table~\ref{tab.roc_curve}. 
\cut{Not surprisingly, four LSE methods perform better than random and they have best performance with Yelp and Twitter datasets.}
\begin{table*}[h]
    \centering   
    \begin{tabular}{ | l | l | l | l |} 
     \hline
     & \textbf{Airbnb} & \textbf{Twitter} & \textbf{Yelp} \\
     \hline
    Increase desirability & closest $\rightarrow$ best & okay $\rightarrow$ good & gorgeous $\rightarrow$ super \\
    or male perception & stores $\rightarrow$ boutiques & sweatheart $\rightarrow$ girlfriend & yummy $\rightarrow$ tasty \\
     & famous $\rightarrow$ old & purse $\rightarrow$ wallet & fabulous $\rightarrow$ excellent \\
     & plaza $\rightarrow$ place & precious $\rightarrow$ rare & hunt $\rightarrow$ search \\
    \hline
    Decrease desirability &  excellent $\rightarrow$ safe & ma $\rightarrow$ mom & tasty $\rightarrow$ yummy\\
     or male perception & best $\rightarrow$ hottest & crib $\rightarrow$ bed & excellent $\rightarrow$ cute\\
     & gorgeous $\rightarrow$ great & impressive $\rightarrow$ wonderful & good $\rightarrow$ yummy\\
     & boutiques $\rightarrow$ stores & buddy $\rightarrow$ boyfriend & attractive $\rightarrow$ cute\\
    \hline
    \end{tabular}
    
    \caption{Word substitutions with high LSE used most frequently by authors of the opposite class (e.g., ``male" words used by female users, and visa versa.) }
    \label{tab.strategy}
\end{table*}

\begin{table}[H]
    \centering
        
    \begin{tabular}{ | c | c | c | c |} 
     \hline
     & \textbf{Yelp} & \textbf{Twitter} & \textbf{Airbnb}  \\
     \hline
    KNN & 0.571 & 0.621 & 0.545  \\		
    VT-RF & 0.707 & {\bf 0.657} & 0.499 \\ 
    CF-RF & 0.617 & 0.613 & {\bf 0.573} \\
    CSF & 0.607 & 0.622 & 0.529  \\
    \hline
    Classifier & {\bf 0.803} & 0.583 & 0.548 \\
    \hline
    \end{tabular}
    \caption{Area under ROC curve}
    \label{tab.roc_curve}
\end{table}

\cut{
\begin{figure*}[h]
	\centering
	\includegraphics[width=0.8\paperwidth]{fig/AMTplot2.png}
	\caption{ ROC curve for treatment effect sign}
	\label{fig:com_cls}
\end{figure*}
}

\subsubsection{Preliminary Analysis using LSE in Online Communication Strategy}
In this section, we provide a preliminary analysis of how LSE estimators may be used to characterize communication strategies online. We show potential communication strategies people use for perception management (try to improve positive perception and reduce negative perception, or to change female style to male or vice versa) according to results suggested by current datasets. 

To do this, we first select top 20 highest and lowest ranked substitutable word pairs according to each LSE estimator. Then, for the 20 highest ranked word pairs, we sort them according to the frequency of positive treatment words used in negative sentence; for the 20 lowest word pairs, we sort them according to the frequency of negative treatment words used in positive sentence. Table \ref{tab.strategy} shows a list of highly ranked word pair selected according to each estimator:

\begin{itemize}
	\item For Airbnb, hosts in undesirable neighborhoods use words {\it best} instead of {\it closest} and {\it boutiques} instead of {\it shops} more often, which are signs of improving desirable perception. While for hosts in desirable neighborhoods, the estimators suggest them to use words {\it excellent} instead of {\it safe} because {\it safe} reduces positive perception compared with {\it excellent} (according to LSE recommendations). This makes sense because hosts located in safe neighborhoods would not emphasize safety.
    \item For gender perception of Twitter and Yelp, LSE estimators recommend that if you want to write sentence like a female, then use {\it sweatheart} instead of {\it girlfriend} and use {\it yummy} instead of {\it tasty}. Otherwise, if you want to write sentences like a male, use {\it buddy} instead of {\it boyfriend} and use {\it attractive} instead of {\it cute}. Additionally, LSE estimators recommend to use more emotional words for female sentence than for male.
\end{itemize}

\newpage

\begin{table*}[!h]
    \centering
    \begin{tabular}{ | p{2cm} | p{10cm} |} 
     \hline
     \multicolumn{2}{| c |}{\textbf{Yelp (make it more likely a female sentence)}}\\
     \hline
     & \\
    Original & Very fresh , and \textit{\underline{tasty}} herbs and spring rolls as well ! \\
     & \\
    KNN/VT-RF/ & Very fresh , and \textit{\underline{yummy}} herbs and spring rolls as well !  \\
    CF-RF/CSF & \\
    \hline
     \multicolumn{2}{| c |}{\textbf{Twitter (make it more likely a male sentence)}}\\
     \hline
     & \\
    Original & Every girl I know is with it and makes \textit{\underline{nice}} dinners for their \textit{\underline{boyfriends}} while I just order pizza and drink \textit{\underline{beer}} with mine \#sorrybabe. \\
    & \\
    KNN/CF-RF & Every girl I know is with it and makes \textit{\underline{good}} dinners for their \textit{\underline{buddies}} while I just order pizza and drink \textit{\underline{beer}} with mine \#sorrybabe.\\
    & \\
    VT-RF/CSF & Every girl I know is with it and makes \textit{\underline{nice}} dinners for their \textit{\underline{buddies}} while I just order pizza and drink \textit{\underline{brew}} with mine \#sorrybabe.\\
     & \\
    \hline
     \multicolumn{2}{| c |}{\textbf{Airbnb (increase desirability)}}\\
    \hline
     & \\
    Original & I don't suggest long walks after dark, but I would \textit{\underline{definitely}} not let this neighborhood discourage your stay, it's \textit{\underline{vibrant}}, fun and \textit{\underline{exciting}}. \\ 
     & \\
    KNN & I don't suggest long walks after dark, but I would \textit{\underline{truely}} not let this neighborhood discourage your stay, it's \textit{\underline{dynamic}}, fun and \textit{\underline{interesting}}.\\ 
    & \\
    VT-RF & I don't suggest long walks after dark, but I would \textit{\underline{really}} not let this neighborhood discourage your stay, it's \textit{\underline{dynamic}}, fun and \textit{\underline{stunning}}.\\
    & \\
    CF-RF/CSF & I don't suggest long walks after dark, but I would \textit{\underline{absolutely}} not let this neighborhood discourage your stay, it's \textit{\underline{dynamic}}, fun and \textit{\underline{spectacular}}.\\
    & \\
    \hline
    \end{tabular}
    
    \caption{Different recommendations of substitution words for one sentence}
    \label{tab.difword_samesents}
\end{table*}

\begin{table*}[t]
    \centering
    \begin{tabular}{ | p{2.2cm} | p{12cm} |} 
     \hline
     \multicolumn{2}{| c |}{\textbf{Yelp (cute $\rightarrow$ attractive)}}\\
    \hline
    & \\
    Largest effect & The joint is \textit{\underline{cute}} and clean and parking is a breeze.\\
    & \\
    Smallest effect & Our \textit{\underline{cute}} Long Island native , Mary suggested the best things on the menu - even telling us what was off and on from the specials board that would work or not.\\
   & \\
    \hline
     \multicolumn{2}{| c |}{\textbf{Twitter (boyfriend $\rightarrow$ buddy)}}\\
     \hline
    & \\
    Largest effect & Monday nights are a night of bonding for me and my \textit{\underline{boyfriend}} ! We both LOVE \#TeenWolf \@user \@user.\\ 
    & \\
    Smallest effect & If you ask me to hang out with you and your \textit{\underline{boyfriend}}  I will look at you like you're stupid then impolitely decline.\\
     & \\
    \hline
     \multicolumn{2}{| c |}{\textbf{Airbnb (store $\rightarrow$ boutique)}}\\
    \hline
    & \\
    Largest effect & Check: Andersonville, in particular, has attracted many gay residents (who have remade the upper reaches of Clark Street into a hot design-\textit{\underline{store}} destination).\\ 
    & \\
    Smallest effect &  Beachwood Village grocery store and coffee \textit{\underline{shop}} conveniently located a mile away.\\
     & \\
    \hline
     
    \end{tabular}
    
    \caption{Sentences that get the largest and smallest treatment effects for a same word pair}
    \label{tab.sameword_difsents}
\end{table*}


\end{document}